\documentclass[runningheads]{llncs}
\usepackage[T1]{fontenc}
\usepackage{graphicx}
\usepackage{booktabs}
\usepackage[misc]{ifsym}

\usepackage{amsmath}
\usepackage{booktabs}
\usepackage{algorithm}
\usepackage{algorithmic}
\usepackage{multirow} 
\usepackage{amsfonts} 
\usepackage{xcolor} 
\usepackage{chngcntr} 
\usepackage{hyperref} 

\newcommand{\corr}{(\Letter)}

\usepackage{eso-pic}
\newcommand{\acceptednotice}{%
  \AddToShipoutPictureFG*{%
    \AtPageLowerLeft{%
      \raisebox{1.7cm}{%
        \makebox[\paperwidth][c]{%
          \footnotesize
          \emph{Accepted at the European Conference on Machine Learning (ECML PKDD) 2026.}%
        }%
      }%
    }%
  }%
}

\begin{document}

\title{Toward Efficient Uncertainty in LLMs \\through Evidential Knowledge Distillation}

\titlerunning{Uncertainty in LLMs through Evidential Knowledge Distillation}

\author{Lakshmana Sri Harsha Nemani\inst{1} \and
P.K. Srijith\inst{1} \and
Tomasz Ku\'smierczyk\inst{2} \corr}

\authorrunning{L.S.H. Nemani et al.}

\institute{Department of Computer Science and Engineering, Indian Institute of Technology Hyderabad, India \and
Jagiellonian University, Faculty of Mathematics and Computer Science, Krak\'ow, Poland\\
\email{tomasz.kusmierczyk@gmail.com}}

\toctitle{Toward Efficient Uncertainty in LLMs through Evidential Knowledge Distillation}
\tocauthor{Lakshmana Sri Harsha Nemani, P.K. Srijith, Tomasz Ku\'smierczyk}

\maketitle              
\acceptednotice         

\begin{abstract}
Accurate uncertainty quantification remains a key challenge for standard LLMs, prompting the adoption of Bayesian and ensemble-based methods. However, such methods typically necessitate computationally expensive sampling, involving multiple forward passes to effectively estimate predictive uncertainty.  
In this paper, we introduce an approach enabling uncertainty estimation in LLMs without incurring the heavy inference latency typically associated with sampling methods. Specifically, we distill uncertainty-aware teachers - originally requiring multiple forward passes - into single-pass students, fine-tuned using LoRA. We compare two distinct distillation strategies: one in which the student employs traditional softmax-based outputs, and another in which the student leverages Dirichlet-distributed outputs to explicitly model epistemic uncertainty via evidential learning.
Empirical evaluation on classification tasks demonstrate that such students can achieve comparable predictive and uncertainty quantification performance relative to their teachers, while requiring only a single forward pass. 

\end{abstract}


\section{Introduction}

Research on uncertainty quantification in LLMs encompasses diverse methodologies~\cite{xia2025survey}, typically categorized into Bayesian-inspired, e.g.,~\cite{ICLR2024_07c256a1}, ensemble-based, e.g.,~\cite{ref1-bayespe}, calibration-based, and verbalization-based approaches.
We focus on the first two, which are theoretically grounded (e.g., account for epistemic uncertainty) and consistently demonstrate improved calibration. However, at the same time, they are computationally intensive and require sampling during both training and inference. The already slow inference of LLMs becomes even slower, as these approaches require multiple forward passes with different parameter samples (Bayesian models) or prompts (model ensembles), effectively limiting their broader adoption.

This work addresses the computational cost of inference by proposing a framework for uncertainty-aware knowledge distillation in text-classifying LLMs.
In particular, we train student models to capture both the predictive performance and the uncertainty estimates of these computationally demanding uncer\-tain\-ty-aware teacher LLMs, while maintaining fast 
inference.
The teacher predictive distributions for different samples are encoded into the student using evidential learning with  Dirichlet distribution over the outputs~\cite{sensoy2018evidentialdeeplearningquantify,ref2-dirich}.
The proposed approach yields student models that provide reliable uncertainty estimates in a single forward pass.

Empirical evaluation across in-distribution and out-of-distribution tasks de\-mon\-strates that our distilled students successfully inherit both the predictive performance and the uncertainty-modeling capabilities of their teachers, while significantly reducing inference latency. 
Although our experiments focus solely on text classification tasks, this work establishes a foundation for developing efficient, uncertainty-aware LLMs for applications beyond text classification.

Due to space limitations, the main text is accompanied by a Supplement, which provides the formal algorithm for our approach, extends the background on sampling-based teachers, and contains details of the experimental setup.
Furthermore, the Supplement includes additional comparison of teachers used in experiments and a study of regularization impact. 
%
The source code of the method has been made publicly available~\footnote{\url{https://github.com/Harsha1969/BPE-KD}}.

\section{Background}

\subsection{Sampling from Teacher Predictive Distributions}
\label{sec:predictive_uncertainty}

We consider two distinct approaches for modeling predictive uncertainty in LLMs. The first approach is based on standard Bayesian Neural Networks (BNNs),  where uncertainty is modeled directly over the  weights~\cite{ritter2018scalable} using Laplace approximation~\cite{ICLR2024_07c256a1}).  The second approach, termed \textit{Bayesian Prompt Ensembles} (BayesPE)~\cite{ref1-bayespe}, employs an ensemble formulation over input prompts, modeling uncertainty arising from prompt variation. 

Both approaches belong to the family of sampling-based models, i.e., they require multiple forward passes to produce predictions. Their predictive distribution is approximated via Monte Carlo (MC) sampling:
\begin{equation}
    p(y \mid x, \mathcal{D}) \approx \sum_{i=1}^{N} w_i \cdot p(y \mid x, \theta_i), \quad \theta_i \sim q(\theta),
    \label{eq:mc_approximation}
\end{equation}
where $y$ denotes the prediction for input $x$, $q(\theta)$ is a distribution encoding uncertainty over a latent variable $\theta$, and $N$ is the number of Monte Carlo samples (or prompt variations).
Each sample $\theta_i$ represents a distinct hypothesis about the latent parameters and is weighted by $w_i$ such that $\sum_{i=1}^{N} w_i = 1$ and $w_i \geq 0$. 
For BNNs, the weights are fixed ($w_i = \frac{1}{N}$), whereas for BayesPE they are learned.
The abstract samples $\theta_i$ correspond to weight samples in Bayesian models or, in the case of prompt ensembles, to varying prompts. 

This sampling procedure translates latent uncertainty regarding the model parameters into predictive uncertainty by implementing Bayesian model averaging. The predictive distribution $p(y \mid x, \mathcal{D})$ defined in Eq.~\eqref{eq:mc_approximation} captures the \textit{total uncertainty} in the model predictions. This uncertainty can be decomposed into two components: \textit{aleatoric uncertainty}, which is inherent to the stochastic nature of the data, and \textit{epistemic uncertainty}, which reflects uncertainty about the model itself.

The entropy $H[Y \mid x, \mathcal{D}]$ of the predictive distribution $p(y \mid x, \mathcal{D})$ quantifies the uncertainty of the model's predictions and, by the \textit{law of total entropy}, can be decomposed~\cite{NEURIPS2018_3ea2db50} as:
$
H[Y \mid x, \mathcal{D}] = \underbrace{\mathbb{E}_{\theta \sim q(\theta)} \left[ H[Y \mid x, \theta] \right]}_{\text{aleatoric}} + \underbrace{I[Y; \theta \mid x, \mathcal{D}]}_{\text{epistemic}},
$
where $H[\cdot]$ denotes Shannon entropy and $I[\cdot; \cdot]$ denotes mutual information. In practice, these quantities are estimated from the MC samples $\{\theta_i\}_{i=1}^{N} \sim q(\theta)$.

\subsection{Fine-tuning Large Models}
\label{sec:lora}

Training large models from scratch is computationally expensive. Instead fine-tuning of a pre-trained model using Low-Rank Adaptation (LoRA) can be used.
LoRA~\cite{ref5-lora} provides a lightweight fine-tuning technique for LLMs by updating a limited number of parameters instead of retraining the entire model. 

Let's consider a frozen weight matrix $W \in \mathbb{R}^{d \times k}$ in a Transformer layer. LoRA augments this matrix with a low-rank update:
$
W_{\text{adapted}} = W + \Delta W = W + BA,
$
where $A \in \mathbb{R}^{r \times k}$ and $B \in \mathbb{R}^{d \times r}$ are trainable low-rank matrices, and $r$ is the adaptation rank, chosen to be much smaller than $d$ and $k$. 
This design allows LoRA to inject task-specific capacity with minimal additional parameters, offering a favorable trade-off between efficiency and flexibility.

The learning process modifies only the parameters in $A$ and $B$, while the original weights $W$ remain unchanged. This approach significantly reduces memory consumption and computational cost. 
To achieve sufficient fine-tuning fidelity, LoRA is typically applied jointly across multiple layers $\{\ell\}$, yielding a set of updates $\{\Delta W_{\ell}\}$ defined through $\{A_{\ell}, B_{\ell}\}$.
LoRA is 
also extended to the final classification layer of the student model.


\section{Method}
The main objective of this work is to enable reliable uncertainty estimation in a single forward pass at inference time. To achieve this, student models need to be designed to approximate both the predictive behavior and the uncertainty estimates of a teacher model while remaining computationally efficient, i.e., emulate the estimates in a single forward-pass.  We achieve this through knowledge distillation and evidential learning based on Dirichlet distribution.  

\subsection{Distillation by Fine-tuning Students}

Distillation is typically performed by training a student model to mimic a teacher. Instead of training LLMs from scratch, which is computationally expensive, we perform \emph{knowledge distillation by fine-tuning a pre-trained model using LoRA} in the way discussed in Section~\ref{sec:lora}. 
%
In particular, if the teacher is already a fine-tuned LoRA model and the student shares the same backbone, we initialize the student by copying the teacher's low-rank matrices, avoiding unnecessary duplication of the base model.
Then, we employ LoRA adapters together with modified (as explained below) classification heads to distill the teacher's predictive distribution.
Note that using the teacher for initialization of the student does not conflict with our objective of fast inference, as the primary cost arises from repeated sampling and not from the cost of a single forward pass.

\subsection{Classification Heads and Training Losses}

We consider two types of student models: the baseline student, which uses the standard categorical (\emph{softmax}) output, and the {evidential student}, which employs a \emph{Dirichlet} distribution. Let $\mathbf{z}(x) = [z_1, \dots, z_K]^\top$ denote the logits from the final classification layer for $K$ classes given an input $x$. Then, the primary difference between the two student types lies in how they parameterize the student's output distribution using these logits.

\subsubsection{Distilling Mean Probabilities with Softmax Outputs}
The baseline student produces a single categorical distribution per input $x$ by mapping logits to probabilities with the \emph{softmax} function: 
$
p_{c}(x) :=
p \left(y=c \mid x\right)
=\sigma \bigl({z}(x)\bigr)_c
=\frac{\exp \bigl(z_c(x)\bigr)}{\sum_{j=1}^K \exp \bigl(z_j(x)\bigr)}.
$
It is trained to approximate the \emph{mean predictive distribution} of a teacher ensemble obtained from sampling 
as given by the Eq.~\eqref{eq:mc_approximation}. 
In particular,
for every input $x^{(i)}$ we have $N$ teacher hypotheses $\{\theta_n\}_{n=1}^N$ with weights $\{w_n\}_{n=1}^N$, 
and
each hypothesis yields a probability vector $p$.
%
The student is then fitted by minimizing the negative log-likelihood of these teacher samples:
 \begin{align}
    \mathcal{L}_{\text{Soft}} 
    &= -\frac{1}{M} \sum_{i=1}^{M} \sum_{n=1}^{N} w_n \sum_{c=1}^{K} p(y=c \mid x^{(i)},\theta_n) \log p_{c}(x^{(i)})  \nonumber \\
 &= -\frac{1}{M} \sum_{i=1}^{M} \sum_{c=1}^{K} \bar{p}_{ \mathcal{T},c}(x^{(i)}) \log p_{c}(x^{(i)}), 
 \end{align}
where $\bar{p}_{ \mathcal{T},c}(x^{(i)}) = \sum_{n=1}^{N} w_n\, p(y=c \mid x^{(i)},\theta_n)$ are teacher's average predictions,
and
$M$ denotes the number of samples in the training dataset.

This approach exhibits certain limitations.
The softmax student can represent only the \emph{mean}
probability vector. 
Figure~\ref{fig:dirichlet} illustrates this learning issue: the blue dots correspond to samples from the teacher's predictive distribution, while the red dot denotes the mean probability vector that the student is trained to approximate.
Multiple distributions can collapse to the same mean. 
Furthermore, while the distilled baseline model aims to reproduce the total predictive uncertainty of the teacher, this uncertainty is collapsed entirely into the Shannon entropy of the categorical distribution
$
H \bigl[Y\mid x,\mathcal{D}\bigr]
= -\sum_{c=1}^{K} p_c(x)
                  \log p_c(x).
$
Consequently, the distinct contributions of aleatoric and epistemic uncertainty are no longer separable.
Thus the softmax student provides fast single-pass predictions at the
cost of discarding higher-order information (variance, covariance) that
is preserved by the Dirichlet student.

\subsubsection{Encoding Predictive Distribution with Dirichlet Outputs}
In a $K$-class classification setting, evidential deep learning models are designed to produce the parameters of a Dirichlet distribution that captures uncertainty over the categorical output space. 
The Dirichlet distribution serves as a fundamental tool for capturing uncertainty about categorical probabilities, representing our beliefs about probability vectors rather than individual outcomes. As a conjugate prior for the categorical distribution, it quantifies the uncertainty inherent in estimating probabilities from limited data.

\begin{figure}[t]
    \centering
    \includegraphics[width=0.7\columnwidth]{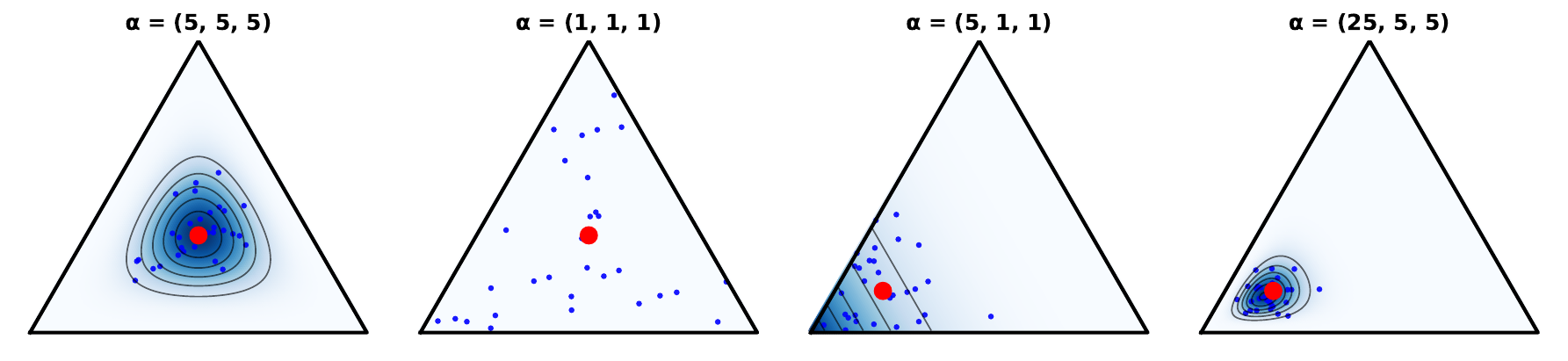}
\caption{
        Dirichlet distributions on the 2-simplex illustrating uncertainty quantification for categorical probabilities. Each panel illustrates different concentration parameters $\alpha$. Blue heatmaps represent the probability densities, red circles indicate the expected probability vectors (means), and blue dots show samples from the distributions - each representing a possible "true" probability vector for the three categories. Despite having the same mean when $\alpha$ values are proportional, the distributions exhibit dramatically different levels of concentration.
}
    \label{fig:dirichlet}
\end{figure}

Rather than outputting class probabilities directly, the neural network predicts a set of concentration parameters $\alpha = [\alpha_1, \dots, \alpha_K]$ corresponding to the parameters of the Dirichlet distribution. 
The parameters are obtained from the logits of the last layer $z_c$ as
$
\alpha_c = 1 + \text{softplus}(z_c)$, 
for $c = 1, \dots, K$,
ensuring that $\alpha_c > 0$.
Then, the class-wise predictive probability for a given class $c$ is computed as $p_c = \frac{\alpha_c}{\alpha_0}$, where $\alpha_0 = \sum_{c=1}^{K} \alpha_c$ represents the total evidence accumulated by the model.
The distribution's concentration parameter 
$\alpha_0$ serves as an explicit measure of prediction confidence. 
It controls the degree of certainty around the expected probabilities, with higher values indicating greater confidence and lower values reflecting increased uncertainty.
A larger value of $\alpha_0$ corresponds to a sharper, more peaked Dirichlet distribution - indicating strong belief in the prediction. Conversely, a smaller $\alpha_0$ suggests greater uncertainty, as it results in a flatter distribution over the classes. This formulation enables the model to generate both class predictions and associated uncertainty estimates in a single deterministic forward pass. Figure~\ref{fig:dirichlet} illustrates this ability to represent both the expected outcome and the confidence in that expectation.

Formally, the Dirichlet distribution over the categorical class probabilities $p = [p_1, \dots, p_K]$ is defined as:
$
\text{Dir}(p \mid \alpha) = \frac{1}{B(\alpha)} \prod_{c=1}^K p_c^{\alpha_c -1},
$
where $B(\alpha)$ is the multivariate Beta function:
$
B(\alpha) = \frac{\prod_{c=1}^K \Gamma(\alpha_c)}{\Gamma(\alpha_0)}$, with $\alpha_0 = \sum_{c=1}^K \alpha_c$.
Taking the log of the density yields the log-likelihood for a single $p$:
$
\log \text{Dir}(p \mid \alpha) = \log \Gamma(\alpha_0) -\sum_{c=1}^K \log \Gamma(\alpha_c) + \sum_{c=1}^K (\alpha_c -1) \log p_c.
$

The Dirichlet student is trained to match the teacher's predictive distribution using a Dirichlet-based distillation loss. For each input $x^{(i)}$ in the training dataset, the student produces Dirichlet concentration parameters $\alpha^{(i)} = [\alpha_1^{(i)}, \dots, \alpha_K^{(i)}]$. On the other hand, the teacher provides samples $p$ (e.g., prompt-wise predictions) along with the weights $w_i$, of the predictive distribution $p(y \mid \theta_n, x^{(i)})$. 
The loss is then the negative log-likelihood of the predictions $p$: \\
 \begin{align}
    \mathcal{L}_\text{Dirichlet} &= -\frac{1}{M} \sum_{i=1}^M [
\log \Gamma(\alpha_0^{(i)}) -\sum_{c=1}^K \log \Gamma(\alpha_c^{(i)})
+ \nonumber
\\&
+\sum_{n=1}^N w_n \sum_{c=1}^K (\alpha_c^{(i)}-1) \cdot \log p(y=c \mid \theta_n, x^{(i)})
].
 \end{align}
It encourages the student's Dirichlet mean to match the teacher's ensemble prediction across prompts, while also learning meaningful uncertainty through the shape of the distribution.

Let $p \sim \mathrm{Dir}(\alpha)$ with total evidence $\alpha_{0} = \sum_{c}\alpha_{c}$.  
The mean predictive probabilities for the student are $\bar p_{c}(x)=\mathbb{E}[p]_c = \alpha_{c}/\alpha_{0}$, whose entropy quantifies the \emph{total} (aleatoric $+$ epistemic) uncertainty:
$
    H[Y \mid x, \mathcal{D}]
    = -\sum_{c=1}^{K}\bar p_{c}(x)\,\log\bar p_{c}(x).
$
Averaging the categorical entropy over the Dirichlet posterior (with $\psi$ the di\-gam\-ma function) yields the \emph{aleatoric} component:
$
    \mathbb{E}_{p} \bigl[ H[Y\mid p] \bigr]
    = -\sum_{c=1}^{K}\frac{\alpha_{c}}{\alpha_{0}}
      \bigl[\psi(\alpha_{c}+1)-\psi(\alpha_{0}+1)\bigr].
$
Their difference
$
    I[Y, p \mid x, \mathcal{D}]
    = H[Y \mid x, \mathcal{D}]
      -\mathbb{E}_{p} \bigl[ H[Y\mid p] \bigr]
$
is the mutual information between label $Y$ and model parameters, measuring \emph{epistemic} uncertainty. Note it vanishes as $\alpha_{0} \to \infty$, i.e., when the model accumulates sufficient evidence.

\subsection{Distillation with Early Stopping}
\label{sec:distill_earlystop}

We distill the teacher $\mathcal{T}$ into a student $\mathcal{S}$ by fine-tuning only a handful of low-rank LoRA adapters and an adapted classification head.  
%
Every mini-batch of training data $(x^{(i)},y^{(i)})$ is passed for \emph{all} $\{\theta_n\}$ to sample the weighted predictive distribution  (as in Eq.~\eqref{eq:mc_approximation}).  The same input $x^{(i)}$ is simultaneously fed to the student, after which the student parameters are updated by back-propagating gradients of appropriate distillation loss $\mathcal{L}$. 
Mini-batch stochastic gradient descent with LoRA adapters ensures that distillation is efficient. 

Similarly,
an early-stopping rule guards against overfitting and unnecessary computation.
Although ground-truth labels are \emph{not} used in the optimization objectives, they are used to compute the per-epoch negative log-likelihood
$
\text{NLL} = -\frac{1}{M}\sum_{i=1}^{M}\log p_{\mathcal{S}}\!\bigl(y^{(i)}\mid x^{(i)}\bigr),
$
where $p_{\mathcal{S}}$ denotes either output probability from the softmax student or   the marginal under the predicted Dirichlet distribution from the evidential student. Monitoring this NLL provides the following model-agnostic criterion for \emph{early stopping}: when the metric does not improve for 5 epochs, training is halted and the best checkpoint is restored. A rising or stagnant NLL indicates the onset of overfitting to teacher noise. In practice, the curve stabilizes after only two to four epochs for BayesPE and after approximately 20-40 epochs for the Bayesian teacher.

The full training procedure is summarized in the Supplement.

\subsection{Uncertainty Regularization}
\label{sec:regularization}

For the evidential student, we employ a network that parameterizes a Dirichlet distribution $\text{Dir}(\boldsymbol{\alpha})$, where the precision $\alpha_0 = \sum_{c=1}^K \alpha_c$ represents the so-called total evidence and is inversely proportional to epistemic uncertainty. Constraining and regularizing this quantity affects the robustness of the model.

A common evidential regularizer minimizes the Kullback-Leibler divergence toward a uniform prior $\text{Dir}(\mathbf{1}_K)$ to effectively push the mean predictive distribution $\mathbb{E}[p^{(i)}]$ for each point $(i)$ toward $1/K$.
Alternatively, regularizing $\alpha^{(i)}_0$ directly constrains the concentration of the distribution without necessarily biasing the predicted class probabilities. While pulling toward $\text{Dir}(\mathbf{1}_K)$ conflates uncertainty with a lack of class-wise preference, the regularization of just $\alpha^{(i)}_0$ allows the student to express high epistemic uncertainty even when the predicted mean is non-uniform, thereby decoupling the evidence scale from the predicted class label.
Consequently, an alternative approach to stabilizing the student is to fix the precisions to a constant value $\alpha^{(i)}_0 = C$. 
In practice, it is
implemented  as renormalization of the values $\alpha_c$ by their sum after each update.

The proposed regularizer prevents the student from becoming overconfident but at the same time deprives it of the ability to express varying levels of epistemic uncertainty across different inputs. Its flexibility is effectively reduced to modeling only aleatoric uncertainty.
To preserve the model's ability to quantify sample-specific uncertainty while preventing evidence over-accumulation, we can instead regularize $\alpha^{(i)}_0$ toward a learnable dataset-wide mean ${\bar{\alpha}_0}$. 
Although the regularizer 
$\mathcal{L}^{(i)}_{reg} = (\alpha^{(i)}_0 -{\bar{\alpha}_0})^2$
can be considered simplistic, as it is derived from the basic prior $\mathcal{N}(\alpha^{(i)}_0 \mid {\bar{\alpha}_0}, 1)$, it anchors the student's precision to the average confidence level for a training dataset, ensuring that the student does not significantly overestimate its certainty while still permitting local deviations. 
One may note that this formulation imposes no explicit constraints on the values of $\alpha^{(i)}_0$, but they remain positive by design, due to the use of softplus.


\section{Experiments}
\label{sec:experiments}


This section presents an empirical study of student models distilled from two distinct teachers: (1) Bayesian Prompt Ensembles (BayesPE)~\cite{ref1-bayespe} and (2) Laplace LoRA~\cite{ICLR2024_07c256a1}.
Note that our primary objective is {reduced computational cost at inference and in terms of performance, the distillation goal is \emph{not} to improve upon the teachers but to match them}, regardless of their absolute performance.
The teachers were selected as sampling-based representatives from their respective model groups and  the evaluation of their performance was conducted in the respective publications by their authors and is not a concern of this work. However, for completeness, we compare them in
the Supplement. 

We utilized Mistral 7B Instruct v0.3~\cite{refe-mistral7b} as the backbone architecture for both students and teachers. This model represents a strong open-weight LLM baseline with competitive performance relative to larger models while remaining computationally tractable for the repeated sampling and distillation required in our study. Its scale 
ensures reproducibility on commonly available research hardware.
Although, we used the same network architectures for both teachers and students to avoid conflating architectural influences with the effects of the distillation process itself, this constraint is not inherent to our approach, and arbitrary network architectures could in principle be used.

Model fine-tuning was conducted using LoRA, and our evaluation focuses on improving inference time and matching classification accuracy, calibration quality, inference latency, and robustness against out-of-distribution (OOD) data. Experiments were performed on four text classification tasks spanning sentiment analysis, topic classification, and social media content: Amazon Reviews Polarity~\cite{ref6-He_2016}, SST2~\cite{ref7-socher-etal-2013-recursive}, Yahoo Answers~\cite{zhang2015character}, and YouTube Comments~\cite{ref9-inproceedings}.


 Additional details regarding our setting can be found in the Supplement.

\subsection{Inference Time Analysis}

The key motivation behind this work is to reduce the inference-time overhead associated with Bayesian and ensemble methods. Since these methods require multiple forward passes (for example, in the case of BayesPE, one per prompt), they become computationally expensive for real-time or large-scale deployments. In contrast, both our student types perform inference in a single forward pass. 

\begin{table}[t!]
\centering
\caption{Inference time: teacher (\emph{BayesPE}) vs. student (\emph{Dirichlet}).}
\label{tab:inference-time}
    \begin{tabular}{l | r r |
    r}
    \hline
    {Dataset} & {BayesPE} [s] & {Dirichlet} [s] & Speed-up\\
    \hline
    Amazon & 4335.59 & {252.45} & 17$\times$\\
    SST2 & 577.39 & {41.68} &  14$\times$\\
    YouTube & 386.47 & {35.26} & 11$\times$ \\
    Yahoo & 9852.25 & {268.40} & 36$\times$ \\
    \hline
    \hline
    \end{tabular}%
\end{table}

To quantify this improvement, 
%
Table~\ref{tab:inference-time} compares the inference times of Ba\-yesPE teacher and the distilled Dirichlet student. 
The student is substantially faster due to its single-pass nature, as the teacher needs to query multiple prompts, which increases inference time.
We omit the exact numbers
for the distilled softmax student as its inference times are almost identical to that of the Dirichlet model. The only difference between them at inference time is the last layer. 

Speed-ups for the Laplace teacher are even more prominent, as hundreds or even thousands of samples (compared to $\sim$10-20 prompts used for BayesPE) may be needed to sufficiently estimate the predictive distribution for the model. However, we have not optimized this hyperparameter and instead used a very conservative value of 10k samples (which was selected to eliminate the impact of a limited number of samples used for distillation and can be further optimized) hence we omit the exact run times for this teacher as noninformative.

\subsection{Performance of Distilled Students}
\label{sec:performance_experiment}

\begin{table*}[t]
    \centering
\caption{
Test data performance of the {teacher} (\emph{BayesPE}) and two distilled {student} models with \emph{Dirichlet} or \emph{Softmax} output layers on four text‑classification datasets.  
Boldface highlights the best value in each dataset.  
For students we report the mean $\pm$ standard deviation 
and the average number of training epochs before early stopping. 
}
\label{tab:results}
\begin{tabular}{ll|l|llll}
\toprule
Dataset & Model & Epoch & Accuracy ($\uparrow$) & ECE ($\downarrow$) & NLL ($\downarrow$) & Brier ($\downarrow$) \\
\midrule
\multirow[t]{5}{*}{Amazon} & BayesPE & - & \textbf{0.959}& 0.021& 0.160& \textbf{0.035}\\
 & Dirichlet & 3.7& \textbf{0.958} ${}_{\pm0.001}$ & \textbf{0.011} ${}_{\pm0.007}$ & \textbf{0.132} ${}_{\pm0.004}$ & \textbf{0.035} ${}_{\pm0.001}$ \\
 & Softmax & 1.0& 0.957 ${}_{\pm0.000}$ & 0.013 ${}_{\pm0.000}$ & 0.138 ${}_{\pm0.001}$ & \textbf{0.034} ${}_{\pm0.000}$ \\
\cline{1-7}
\multirow[t]{5}{*}{SST2} & BayesPE & - & \textbf{0.955}& 0.029& 0.165& \textbf{0.037}\\
 & Dirichlet & 4.3& \textbf{0.954} ${}_{\pm0.000}$ & \textbf{0.017} ${}_{\pm0.005}$ & \textbf{0.142} ${}_{\pm0.009}$ & \textbf{0.037} ${}_{\pm0.001}$ \\
 & Softmax & 1.3& 0.952 ${}_{\pm0.001}$ & 0.025 ${}_{\pm0.001}$ & 0.147 ${}_{\pm0.004}$ & 0.037 ${}_{\pm0.000}$ \\
\cline{1-7}
\multirow[t]{5}{*}{YouTube} & BayesPE & - & 0.875& 0.031& 0.295& 0.091\\
 & Dirichlet & 4.3& \textbf{0.900} ${}_{\pm0.017}$ & 0.097 ${}_{\pm0.023}$ & 0.294 ${}_{\pm0.022}$ & \textbf{0.079} ${}_{\pm0.010}$ \\
 & Softmax & 4.7& 0.892 ${}_{\pm0.001}$ & \textbf{0.015} ${}_{\pm0.006}$ & \textbf{0.279} ${}_{\pm0.006}$ & 0.084 ${}_{\pm0.002}$ \\
\cline{1-7}
\multirow[t]{5}{*}{Yahoo} & BayesPE & - & 0.593& 0.194& 2.173& 0.061\\
 & Dirichlet & 1.0& \textbf{0.610} ${}_{\pm0.003}$ & \textbf{0.042} ${}_{\pm0.001}$ & \textbf{1.385} ${}_{\pm0.008}$ & \textbf{0.055} ${}_{\pm0.000}$ \\
 & Softmax & 1.0& \textbf{0.609} ${}_{\pm0.003}$ & 0.123 ${}_{\pm0.071}$ & 1.780 ${}_{\pm0.345}$ & 0.057 ${}_{\pm0.002}$ \\
\cline{1-7}
\bottomrule
\end{tabular}
\end{table*}

We evaluate the effectiveness of our approach by quantifying how closely student models approximate strong teachers in 
both accuracy and calibration metrics. 

Table \ref{tab:results} shows performance for two students: the conventional {Softmax} and the evidential student with {Dirichlet} output,
compared to the BayesPE teacher.
Across all benchmarks the distilled students match (or sometimes even exceed) the teacher. The Dirichlet variant attains parity in accuracy on Amazon~($0.958$ vs $0.959$) and SST2 ($0.954$ vs $0.955$) and overtakes the teacher on Yahoo~($+1.7$ pp) and YouTube~($+2.5$ pp). It simultaneously delivers the best likelihoods and lowest Brier scores everywhere, and halves the teacher's ECE on Amazon, SST2 and even more on Yahoo.
The Softmax student is competitive but loses to Dirichlet on all datasets except for YouTube, where it gets very low ECE, but at the cost of slightly higher Brier and lower accuracy. 

Interestingly, the students' improved accuracy and uncertainty quantification cannot arise from knowledge transfer alone. Theoretically, assuming perfect distillation, the students have no means to surpass the performance of the teacher. The observed outcomes appear to be a byproduct of the distillation process (e.g., model selection via early stopping) and model architectures. In particular, by comparing against the Softmax student, we can partially attribute some of these improvements to the regularizing effect of the Dirichlet distribution.

\begin{table*}[t]
    \centering
\caption{
Test data performance of the \emph{teacher} (Laplace) and two distilled \emph{student} models with Dirichlet or Softmax output layers on four text‑classification datasets.
Metrics: accuracy ($\uparrow$), expected calibration error (ECE $\downarrow$), negative log‑likelihood (NLL $\downarrow$), and Brier score ($\downarrow$); boldface highlights the best value in each dataset.
For students we report the mean $\pm$ standard deviation 
and the average number of training epochs before early stopping.
}
\label{tab:laplace_results}
    \begin{tabular}{ll|r|llll}
    \toprule
     &  & Epoch & Accuracy & ECE & NLL & Brier \\
    Dataset & Model &  &  &  &  &  \\
    \midrule
    \multirow[t]{3}{*}{Amazon} & Laplace & - & \textbf{0.973} & \textbf{0.009} & \textbf{0.092} & \textbf{0.023} \\
     & Dirichlet & 43.0 & 0.966 ${}_{\pm0.001}$ & \textbf{0.010} ${}_{\pm0.002}$ & 0.105 ${}_{\pm0.002}$ & 0.026 ${}_{\pm0.000}$ \\
     & Softmax & 23.0 & 0.965 ${}_{\pm0.000}$ & 0.013 ${}_{\pm0.002}$ & 0.107 ${}_{\pm0.003}$ & 0.027 ${}_{\pm0.001}$ \\
    \cline{1-7}
    \multirow[t]{3}{*}{SST2} & Laplace & - & 0.961 & 0.020 & 0.125 & 0.033 \\
     & Dirichlet & 23.0 & 0.961 ${}_{\pm0.002}$ & 0.019 ${}_{\pm0.006}$ & 0.125 ${}_{\pm0.004}$ & 0.033 ${}_{\pm0.002}$ \\
     & Softmax & 37.0 & \textbf{0.964} ${}_{\pm0.002}$ & \textbf{0.010} ${}_{\pm0.004}$ & \textbf{0.118} ${}_{\pm0.004}$ & \textbf{0.029} ${}_{\pm0.001}$ \\
    \cline{1-7}
    \multirow[t]{3}{*}{YouTube} & Laplace & - & 0.932 & \textbf{0.015} & \textbf{0.170} & \textbf{0.050} \\
     & Dirichlet & 15.0 & \textbf{0.932} ${}_{\pm0.008}$ & \textbf{0.021} ${}_{\pm0.005}$ & \textbf{0.176} ${}_{\pm0.008}$ & \textbf{0.050} ${}_{\pm0.002}$ \\
     & Softmax & 14.0 & 0.929 ${}_{\pm0.009}$ & 0.019 ${}_{\pm0.003}$ & 0.182 ${}_{\pm0.007}$ & 0.053 ${}_{\pm0.002}$ \\
    \cline{1-7}
    \multirow[t]{3}{*}{Yahoo} & Laplace & - & 0.717 & 0.022 & 0.823 & 0.040 \\
     & Dirichlet & 46.0 & \textbf{0.732} ${}_{\pm0.001}$ & \textbf{0.012} ${}_{\pm0.003}$ & 0.788 ${}_{\pm0.001}$ & \textbf{0.038} ${}_{\pm0.000}$ \\
     & Softmax & 41.0 & 0.731 ${}_{\pm0.001}$ & 0.009 ${}_{\pm0.003}$ & \textbf{0.772} ${}_{\pm0.001}$ & \textbf{0.038} ${}_{\pm0.000}$ \\
    \cline{1-7}
    \bottomrule
    \end{tabular}
\end{table*}

The results for the Laplace teacher distillation, presented in Table~\ref{tab:laplace_results}, reveal similar patterns, with student models approaching and occasionally (but less frequently) exceeding the performance of their more complex teacher. On the SST2, YouTube Comments, and Amazon Reviews datasets, the Dirichlet students achieve near-parity with the Laplace teacher in terms of both uncertainty scores (NLL and Brier) and accuracy (e.g., exact match on YouTube). 
On the Yahoo Answers dataset, both students outperform the teacher. Notably, the Dirichlet student improves accuracy by $1.5$ percentage points ($0.732$ vs. $0.717$) and reduces ECE from $0.022$ to $0.012$. We also observe highly competitive results for the Softmax student on the SST2 dataset, where it outperforms both the Dirichlet student and the Laplace teacher (both perform almost identically).

In both Table~\ref{tab:results} and Table~\ref{tab:laplace_results}, 
we record also the mean number of gradient-descent epochs required until early stopping. 
For the BayesPE teacher, students converge  in one epoch on Yahoo
and in no more than few epochs on the remaining ones.
This demonstrates that uncertainty information from BayesPE can be transferred without much optimisation overhead. 
On the other hand, for the Laplace teacher, both students required significantly more epochs to converge (around $20-40$). This suggests a more challenging optimization landscape for the Laplace teacher compared to BayesPE.

\begin{table*}[t]
\caption{    
    OOD detection performance 
    for the BayesPE teacher and distilled students
    for YouTube/SST2/Yahoo (OOD) vs Amazon (in-domain; ID) data. 
    Figure~\ref{fig:ood_uncertaintites} illustrates the uncertainty distributions used for these summaries.
}
\label{tab:predictive_entropy}
\centering
     \begin{tabular}{ll|rrr|rrr|rrr}
    \toprule
     &  & \multicolumn{3}{c}{Total} & \multicolumn{3}{c}{Epistemic} & \multicolumn{3}{c}{Aleatoric} \\
    dataset & model  & Avg & $W_1$ & AUROC & Avg & $W_1$ & AUROC & Avg & $W_1$ & AUROC \\


    \midrule
    \multirow[t]{3}{*}{SST2} & BayesPE & 0.05 & 0.01 & 0.40 & 0.02 & 0.01 & 0.47 & 0.03 & 0.01 & 0.39 \\
     & Dirichlet & 0.13 & 0.04 & \textbf{0.82} & 0.02 & 0.01 & \textbf{0.82} & 0.11 & 0.03 & \textbf{0.82} \\
     & Softmax & 0.05 & 0.01 & 0.45 &  -  &  -  &  -  &  -  &  -  &  -  \\
    
    \cline{1-11}
    \multirow[t]{3}{*}{YouTube} & BayesPE & 0.25 & 0.19 & 0.86 & 0.08 & 0.07 & 0.89 & 0.17 & 0.12 & 0.85 \\
     & Dirichlet & 0.60 & 0.51 & \textbf{0.96} & 0.03 & 0.03 & \textbf{0.90} & 0.57 & 0.49 & \textbf{0.99} \\
     & Softmax &  -  & 0.23 & 0.92 &  -  &  -  &  -  &  -  &  -  &  -  \\
    \cline{1-11}
    \multirow[t]{3}{*}{Yahoo} & BayesPE & 0.52 & 0.47 & 0.90 & 0.13 & 0.12 & 0.92 & 0.40 & 0.35 & 0.90 \\
     & Dirichlet & 2.16 & 2.06 & \textbf{1.00} & 0.08 & 0.07 & \textbf{0.95} & 2.07 & 1.99 & \textbf{1.00} \\
     & Softmax &  -  & 0.44 & 0.83 &  -  &  -  &  -  &  -  &  -  &  -  \\
    \cline{1-11}
\bottomrule
    \end{tabular}
\end{table*}

\subsection{Out-of-Distribution Uncertainty Estimation}

We evaluate how well the models quantify uncertainty under distribution shifts through an OOD generalization experiment. All models were trained  exclusively (or distilled from teachers trained) on the Amazon training data and subsequently assessed on test data from the four benchmarks.

\begin{figure}[ht!]
    \centering
    \includegraphics[width=0.825\linewidth]{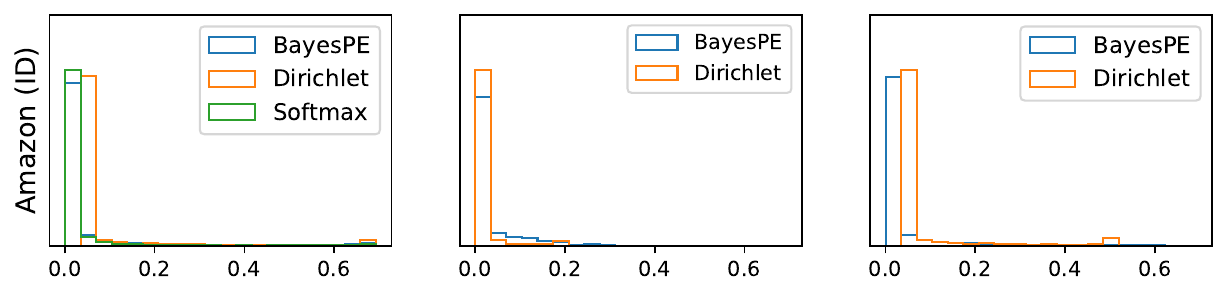}
    \includegraphics[width=0.825\linewidth]{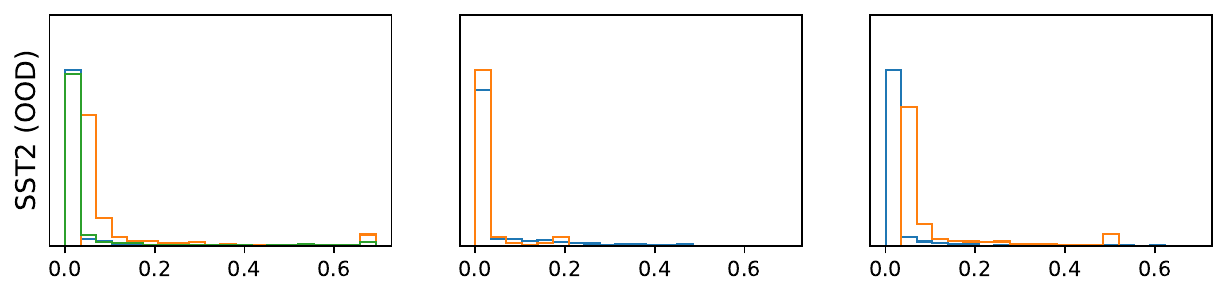}
    \includegraphics[width=0.825\linewidth]{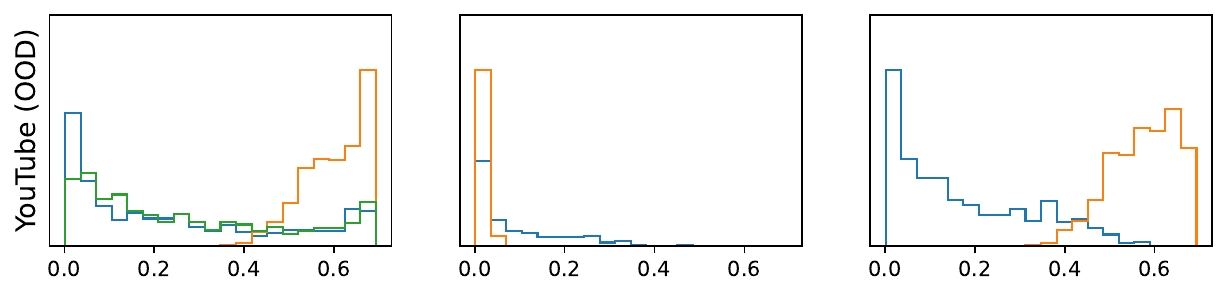}
    \includegraphics[width=0.825\linewidth]{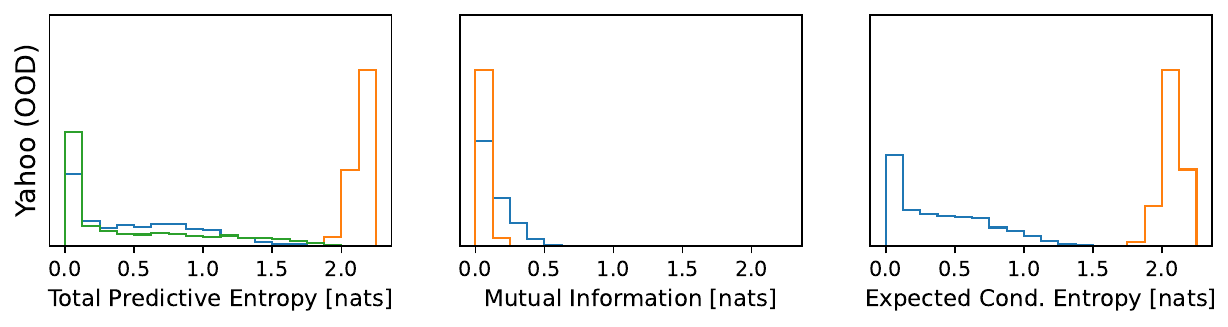}
    \caption{
    Predictive uncertainty distributions for the BayesPE teacher \textbf{(blue)}, Dirichlet student \textbf{(orange)} and Softmax student \textbf{(green)}. Each row shows the empirical distributions of \textbf{(left):} total predictive entropy, \textbf{(middle):} mutual information (epistemic uncertainty), and \textbf{(right):} expected conditional entropy (aleatoric uncertainty) on in-domain Amazon reviews (Row 1) and three OOD datasets: SST2 (Row 2), YouTube (Row 3) and Yahoo (Row 4).
    The last row (Yahoo), has a different number of classes (10 vs. 2) and therefore entropies for it fall into a different range.
    }
    \label{fig:ood_uncertaintites}
\end{figure}

Table~\ref{tab:predictive_entropy} summarizes  predictive entropies decomposed into epistemic and alea\-toric uncertainties for students distilled from the BayesPE  teacher.
Besides the uncertainty averages, we report also 
Wasserstein-1 distance~\cite{villani2008optimal} ($W_1$) and AUROC~\cite{FAWCETT2006861}. $W_1$ measures the discrepancy between in-domain (Amazon) and OOD entropy distributions, while AUROC measures the model's discriminative capability between in-domain and OOD samples. Figure~\ref{fig:ood_uncertaintites} further illustrates these uncertainty distributions through histograms.

The Dirichlet-based student demonstrates a strong but context-dependent capacity for OOD detection. 
It consistently outperforms both the teacher and the Softmax baseline. Compared to the teacher, it amplifies total predictive entropy across all datasets, which translates to superior discrimination and high AUROC scores. 
However, we observe that while the distilled Dirichlet student increases both aleatoric and epistemic uncertainty, the gains are slightly more  pronounced for the former. 

The student being able to quantify uncertainty better than the teacher can be explained only as a side effect of 
the distillation process with early stopping
or of 
using a model with a certain structure, i.e.,
structural biases due to regularizing effect of the Dirichlet distribution. 
Ultimately,
the results are compatible with our conclusions from Section~\ref{sec:performance_experiment}.
%

%

\subsection{Regularization Impact}

\begin{table*}[t]
\centering
\caption{
Comparison of regularization strategies for the Dirichlet student (with Laplace teacher) across four text classification datasets. We evaluate the impact of different precision constraints on model calibration (ECE, NLL, Brier) and predictive accuracy. Metrics for "no regularization" (the default Dirichlet student) represent the mean and standard deviation across multiple runs, while other variants show a single representative run.
}
\label{tab:regularization_results}
\resizebox{1.0\textwidth}{!}{
    \begin{tabular}{ll|r|llll}
    \toprule
    Dataset & Regularization & Epoch & Accuracy ($\uparrow$) & ECE ($\downarrow$) & NLL ($\downarrow$) & Brier ($\downarrow$)\\
    \midrule
    \multirow[t]{5}{*}{Amazon} & no regularization & 42.7 & \textbf{0.966} ${}_{\pm0.001}$ & \textbf{0.010} ${}_{\pm0.002}$ & \textbf{0.105} ${}_{\pm0.002}$ & \textbf{0.026} ${}_{\pm0}$ \\
     & Dir($1_K$) prior & 33& 0.967 & 0.053 & 0.134 & 0.029 \\
     & $\alpha_0$=5K (concentrated) & 36& 0.966 & 0.018 & 0.116 & 0.028 \\
     & $\alpha_0$=K (matching) & 24& 0.965 & 0.077 & 0.159 & 0.032 \\
     & pulling to $\bar{\alpha}_0$ & 50& 0.965 & 0.012 & 0.128 & 0.030 \\
    \cline{1-7}
    \multirow[t]{5}{*}{SST2} & no regularization & 23.3 & 0.961 ${}_{\pm0.002}$ & 0.019 ${}_{\pm0.006}$ & 0.125 ${}_{\pm0.004}$ & 0.033 ${}_{\pm0.002}$ \\
     & Dir($1_K$) prior & 27& 0.962 & 0.067 & 0.156 & 0.037 \\
     & $\alpha_0$=5K (concentrated) & 31& \textbf{0.969} & \textbf{0.003} & \textbf{0.123} & \textbf{0.029} \\
     & $\alpha_0$=K (matching) & 28& 0.960 & 0.071 & 0.168 & 0.037 \\
     & pulling to $\bar{\alpha}_0$ & 49& 0.961 & 0.029 & 0.151 & 0.034 \\
    \cline{1-7}
    \multirow[t]{5}{*}{Yahoo} & no regularization & 46& \textbf{0.732} ${}_{\pm0.001}$ & \textbf{0.012} ${}_{\pm0.003}$ & \textbf{0.788} ${}_{\pm0.001}$ & \textbf{0.038} ${}_{\pm0}$ \\
     & Dir($1_K$) prior & 50& 0.732 & 0.157 & 0.937 & 0.041 \\
     & $\alpha_0$=5K (concentrated) & 50& 0.730 & 0.031 & 0.791 & 0.038 \\
     & $\alpha_0$=K (matching) & 28& 0.731 & 0.148 & 0.893 & 0.041 \\
     & pulling to $\bar{\alpha}_0$ & 49& 0.727 & 0.051 & 0.824 & 0.039 \\
    \cline{1-7}
    \multirow[t]{5}{*}{YouTube} & no regularization & 15.3 & 0.932 ${}_{\pm0.008}$ & 0.021 ${}_{\pm0.005}$ & 0.176 ${}_{\pm0.008}$ & 0.050 ${}_{\pm0.002}$ \\
     & Dir($1_K$) prior & 26& 0.927 & 0.111 & 0.260 & 0.069 \\
     & $\alpha_0$=5K (concentrated) & 16& \textbf{0.944} & \textbf{0.018} & \textbf{0.166} & \textbf{0.045} \\
     & $\alpha_0$=K (matching) & 13& 0.940 & 0.077 & 0.216 & 0.053 \\
     & pulling to $\bar{\alpha}_0$ & 41& 0.931 & 0.065 & 0.245 & 0.061 \\
    \cline{1-7}
    \bottomrule
    \end{tabular}
}
\end{table*}

We examine the impact of regularization of the  concentration parameter $\alpha_0$ on uncertainty  by comparing the strategies introduced in Section~\ref{sec:regularization} against the no-regularization default Dirichlet student. The results in Table~\ref{tab:regularization_results} reveal dominance of the two leading approaches: the non-regularized baseline and the Dirichlet student with 'concentrated'fixed precision $\alpha_0 = 5K$ (where $K$ denotes the number of classes). 
In particular, the non-regularized approach wins in all metrics on the Amazon and Yahoo datasets, while the approach with $\alpha_0 = 5K$ on SST2 and YouTube. On the other hand, the stronger regularizers such as $\alpha_0 = K$ and $\text{Dir}(\textbf{1}_K)$ consistently degrade ECE and NLL. Finally, while our proposed $\bar{\alpha}_0$ regularization does not outperform the best approaches, it maintains competitive performance across all benchmarks.

The Supplement allows us
to dive deeper into the effect of using the fixed global $\alpha_0=5K$ versus learning the sample-specific $\alpha_0$. 
It illustrates distributions of the performance and uncertainty metrics on the YouTube Comments test set. 
Let's recall that for the BayesPE teacher, it was the only data set for which the Dirichlet student struggled to quantify uncertainty.
The plots confirm that, while learning individual $\alpha_0$ per sample achieves consistently strong calibration and likelihood, it is sometimes possible to  improve upon these results by selecting an appropriate fixed global $\alpha_0$.

An extended discussion is provided in the Supplement. 

\subsection{Prompt Impact}

We analyze the impact of prompt quality on the performance and calibration of distilled student models, guided by prompt weights from the BayesPE teacher.

\begin{table*}[t]
\centering
\caption{
Test data performance of the teacher model (BayesPE) and the distilled students for prompts of varying fit quality (as measured by the weight $w_i$). 
}
\label{tab:prompt}
\resizebox{1.0\textwidth}{!}{
    \begin{tabular}{l|ll|r|llll}
    \toprule
 Dataset & Model & Prompt & Epoch & Accuracy ($\uparrow$) & ECE ($\downarrow$) & NLL ($\downarrow$) & Brier ($\downarrow$) \\
    \midrule
    \multirow[t]{7}{*}{SST2} & BayesPE &   &  - & 0.955 & 0.029 & 0.165 & 0.037 \\
    \cline{2-8}
     & \multirow[t]{3}{*}{Dirichlet}  
     & best & 4.3 & 0.954 ${}_{\pm0.000}$ & 0.017 ${}_{\pm0.005}$ & 0.142 ${}_{\pm0.009}$ & 0.037 ${}_{\pm0.001}$ \\
     & 
     & average & 5& 0.957 ${}_{\pm0.003}$ & 0.020 ${}_{\pm0.005}$ & 0.135 ${}_{\pm0.004}$ & 0.036 ${}_{\pm0.001}$ \\
     & 
     & worst & 4& 0.956 ${}_{\pm0.004}$ & 0.018 ${}_{\pm0.002}$ & 0.136 ${}_{\pm0.005}$ & 0.036 ${}_{\pm0.001}$ \\
    \cline{2-8}
     & \multirow[t]{3}{*}{Softmax}  
     & best & 1.3 & 0.952 ${}_{\pm0.001}$ & 0.025 ${}_{\pm0.001}$ & 0.147 ${}_{\pm0.004}$ & 0.037 ${}_{\pm0.000}$ \\
     & 
     & average & 2& 0.958 ${}_{\pm0.000}$ & 0.023 ${}_{\pm0.000}$ & 0.147 ${}_{\pm0.001}$ & 0.036 ${}_{\pm0.000}$ \\
     & 
     & worst & 1& 0.956 ${}_{\pm0.001}$ & 0.026 ${}_{\pm0.001}$ & 0.154 ${}_{\pm0.001}$ & 0.037 ${}_{\pm0.000}$ \\
    \cline{1-8} \cline{2-8}
    \multirow[t]{7}{*}{YouTube Comments} & BayesPE &  & - & 0.875 & 0.031 & 0.295 & 0.091 \\
    \cline{2-8}
     & \multirow[t]{3}{*}{Dirichlet}  
     & best & 4.3 & 0.900 ${}_{\pm0.017}$ & 0.097 ${}_{\pm0.023}$ & 0.294 ${}_{\pm0.022}$ & 0.079 ${}_{\pm0.010}$ \\
     & 
     & average & 8.7 & 0.890 ${}_{\pm0.009}$ & 0.048 ${}_{\pm0.020}$ & 0.277 ${}_{\pm0.020}$ & 0.081 ${}_{\pm0.004}$ \\
     & 
     & worst & 5.7 & 0.906 ${}_{\pm0.004}$ & 0.079 ${}_{\pm0.022}$ & 0.281 ${}_{\pm0.015}$ & 0.078 ${}_{\pm0.002}$ \\
    \cline{2-8}
     & \multirow[t]{3}{*}{Softmax}  
     & best & 4.7 & 0.892 ${}_{\pm0.001}$ & 0.015 ${}_{\pm0.006}$ & 0.279 ${}_{\pm0.006}$ & 0.084 ${}_{\pm0.002}$ \\
     & 
     & average & 1& 0.875 ${}_{\pm0.001}$ & 0.035 ${}_{\pm0.004}$ & 0.288 ${}_{\pm0.002}$ & 0.086 ${}_{\pm0.001}$ \\
     & 
     & worst & 1& 0.933 ${}_{\pm0.001}$ & 0.022 ${}_{\pm0.000}$ & 0.196 ${}_{\pm0.001}$ & 0.055 ${}_{\pm0.000}$ \\
    \cline{1-8} \cline{2-8}
    \bottomrule
    \end{tabular}
}
\end{table*}

Results for best, average, and worst prompts are summarized in Table~\ref{tab:prompt}. 
These categories are based directly on the prompt weights $w_n$ learned by the BayesPE teacher during its training phase, where weight variability reflects how strongly the teacher relied on specific prompts to form its ensemble prediction.
For the SST2 dataset, we observe minimal variation across prompt choices. Accuracy, calibration (ECE), negative log-likelihood (NLL), and Brier scores remain consistently robust, suggesting that prompt selection has negligible influence in this context. On the other hand, for the YouTube Comments dataset (which is the only dataset (see Table~\ref{tab:results}) where the Dirichlet student exhibited worse calibration than both the teacher and the competing Softmax student), prompt quality significantly impacts calibration metrics. Specifically, for the Dirichlet student, ECE scores vary notably, with the best prompt (ECE $=0.097$) exhibiting worse calibration than average (ECE $=0.048$) and the worst prompts (ECE $=0.079$). This variability underscores the sensitivity of calibration to prompt selection in datasets where the student model inherently demonstrates weaker calibration compared to the teacher model.

\section{Related Work}

This work focuses on the distillation of sampling-based LLMs. We discuss the most relevant related work in Section~\ref{sec:uq_distillation}. Although our objective is \emph{not} to introduce a new approach for modeling uncertainty, we briefly review uncertainty modeling in LLMs in Section~\ref{sec:related_uq} to situate sampling-based models within the broader research landscape. Finally, we also cover approaches aimed at improving the efficiency of BNNs (but unlike us, outside the LLM context) through architectural modifications or learning trade-offs, such as the work of~\cite{harrison2024variational}.

\subsection{Uncertainty Modeling in LLMs}
\label{sec:related_uq}
Diverse methodologies for uncertainty modeling in LLMs can be categorized~\cite{xia2025survey} into Bayesian-, ensemble-, calibration-, and verbalization-based approaches.

The first two categories belong to the family of sampling approaches and are of primary interest to our study. 
Bayesian-inspired methods treat model parameters as random variables and leverage approximate inference, e.g., via MC Dropout, Variational or Laplace approximations over weights or LoRA weights~\cite{ICLR2024_07c256a1,onal2024gaussian,NEURIPS2024_7d535754,shi2025trainingfree} to handle epistemic uncertainty. Ensemble-based methods, including prompt ensembles such as BayesPE~\cite{ref1-bayespe} and other perturbation-based approaches, e.g., SPUQ~\cite{gao-etal-2024-spuq}, aggregate multiple model predictions to enhance robustness and uncertainty quantification.
Another method of this type could be the framework proposed by~\cite{10.5555/3692070.3692835} for high-fidelity uncertainty decomposition by ensembling multiple input clarifications.
We emphasize that any of these methods can serve as an example of a computationally demanding teacher.

On the other hand,
calibration and post-hoc methods are 
outside of our scope,
as they operate by directly adjusting predicted (single-pass) probabilities
to better reflect observed accuracy. 
Examples there include temperature scaling and length-invariant normalization, e.g.,~\cite{vashurin2025uncertaintyline}. Similarly, verbalization-based methods use explicit linguistic signals of uncertainty produced by the model itself, effectively capturing uncertainty in tasks requiring nuanced reasoning, e.g.,~\cite{tao2025revisiting}.

\subsection{Knowledge and Uncertainty Distillation}
\label{sec:uq_distillation}
Knowledge Distillation (KD)~\cite{ref4-KDNN}  is a popular approach for compressing large models by training a student model, generally simpler and smaller, to replicate the behavior of a more complex, pre-trained teacher. KD has also been shown to be effective for transferring uncertainty information from teacher to student.

A recent work by \cite{ref3-BKD} proposed Bayesian KD, which grounded conventional KD within the Bayesian framework. In this approach, the teacher model's output is treated as a prior for the student model,
and a posterior is found using Stochastic Gradient Langevin Dynamics~\cite{sgld}.
The distillation there transfers knowledge between models in a Bayesian way, but does not help to improve computational overhead, which is the primary goal here.


Evidential Deep Learning (EDL) is an increasingly popular, efficient alternative to sampling-based uncertainty estimation~\cite{sensoy2018evidentialdeeplearningquantify}. 
For example,
one recent study~\cite{li2025calibratingllmsinformationtheoreticevidential} improves model reliability by incorporating an information bottleneck into Evidential Deep Learning to prevent the model from becoming overconfident. While they focuse on improving the training process of a single model, our approach employs evidential learning within a knowledge distillation framework. Rather than regularizing a single model, we aim to transfer complex uncertainty information from a computationally expensive teacher model to a significantly faster student model.
Similar to us, \cite{ref2-dirich} explored integrating BNNs and EDL via KD, using a computationally intensive BNN teacher to guide a Dirichlet-based student. While our approach also utilizes Dirichlet outputs, it differs in three key areas. First, we focus on fine-tuning LLMs with LoRA rather than training standard neural networks. Second, our training objective is the KL divergence solely on soft teacher labels, whereas \cite{ref2-dirich} trains primarily with ground-truth labels. Third, they achieve calibration through a post-processing refinement with NFs to relax Dirichlet conjugate prior assumptions. In contrast, we utilize an optimization strategy with early stopping based on training data NLL.

A parallel effort by \cite{vejendla2025efficient} seeks to eliminate the multi-pass overhead of Bayesian LLMs through distillation, similar to our work. However, their approach is limited to standard Bayesian teachers and standard softmax students. In contrast, our framework also explores an alternative paradigm using evidential KD. We employ different approaches for training, loss functions, and regularization, while providing a deeper dive into uncertainty quantification. Finally, unlike their white-box method, our approach is compatible also with black-box teachers.

\section{Conclusion}

In this paper,
we introduced a distillation framework for transferring uncertainty from sampling-based (Bayesian or ensemble) LLM teachers to lightweight, single-pass student models. 
In particular,
we studied
both the standard Softmax students and evidential students with Dirichlet-distributed outputs and additional uncertainty regularization. Softmax students can approximate the teacher's predictive mean but are limited to modeling aleatoric uncertainty, whereas Dirichlet students may retain higher-order uncertainty structure and explicitly support epistemic uncertainty modeling.
To our knowledge, this is the first application of Evidential Deep Learning to the distillation of Large Language Models.
While currently limited to discrete text classification, our work serves as a proof-of-concept for single-pass uncertainty estimation, which future work could extend to broader contexts such as open text generation.

Across multiple text classification benchmarks, distilled students achieve predictive accuracy and uncertainty estimates comparable to those of the teachers, while achieving the substantially reduced inference time.
In our experiments, Dirichlet students offered more balanced improvements in calibration and generalization. 
Their inductive biases were particularly pronounced when distilling from the BayesPE teacher, for which the Dirichlet student yielded lower calibration error and more stable uncertainty estimates. 
On the other hand, Softmax students, while lacking explicit epistemic uncertainty modeling, remained competitive (especially under Laplace teacher), suggesting that when the teacher's predictive distribution is more complex, a simpler parameterization can approximate it more efficiently by avoiding optimization difficulties.
Overall, we conclude that Dirichlet students offer more principled modeling characteristics, whereas Softmax students provide a robust alternative with solid performance in less uncertainty-sensitive regimes.

\section*{Acknowledgments}

This research is part of the project No. {2022/45/P/ST6/02969} co-funded by the National
Science Centre and the European Union Framework Programme for Research and
Innovation Horizon 2020 under the Marie Skłodowska-Curie grant agreement No.
945339. For the purpose of Open Access, the author has applied a CC-BY public copyright
licence to any Author Accepted Manuscript (AAM) version arising from this submission. 
\\
\includegraphics[width=1cm]{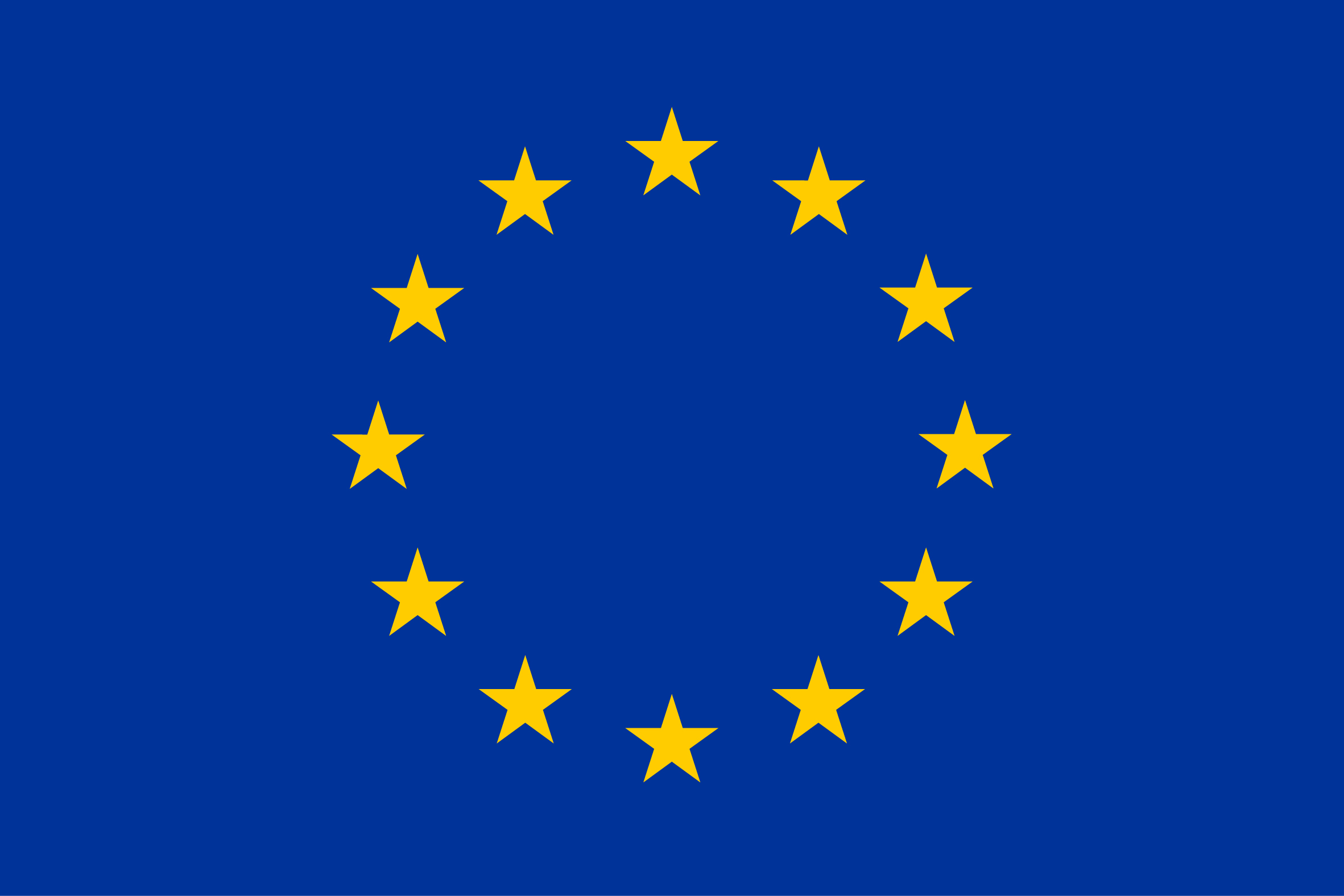} \includegraphics[width=1.9cm]{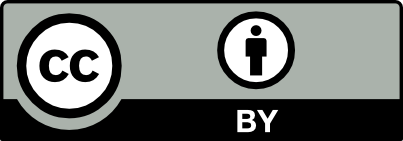}
\\ \\
We gratefully acknowledge Polish high-performance computing infrastructure PLGrid (HPC Center: ACK Cyfronet AGH) for providing computer facilities and support within the computational grant no. PLG/2024/017893. We also gratefully acknowledge Japan International Cooperation Agency (JICA) for the high performance computing facility at IIT Hyderabad.

\bibliographystyle{splncs04}
\bibliography{references}

\clearpage


\begin{center}
{\Large\bfseries Supplementary Material}\\[4pt]
{\large Toward Efficient Uncertainty in LLMs through Evidential Knowledge Distillation}
\end{center}
\vspace{1.5em}

\appendix


\newcommand{\setupSupplement}{
    \setcounter{table}{0}
    \setcounter{figure}{0}
    
    \renewcommand{\thetable}{\thesection.\arabic{table}}
    \renewcommand{\thefigure}{\thesection.\arabic{figure}}
}

\counterwithin{figure}{section}
\counterwithin{table}{section}

\hfill
\hfill

\section{Training Algorithm}

\begin{algorithm}[ht]
\caption{Training the student LLM via distillation}
\begin{algorithmic}[1]
\REQUIRE Teacher model $\mathcal{T}$, student model $\mathcal{S}$ (with LoRA), training data $\mathcal{D}_{\text{train}} = \{(x^{(i)}, y^{(i)})\}_{i=1}^M$, samples/prompts $\{\theta_1, \dots, \theta_N\}$, weights $\{w_1, \dots, w_N\}$ 
\STATE Let $Q_{\text{best}}$ be a prompt (e.g., for BayesPE we pick the one with the highest weight $w_n$)
\FOR{epoch = 1 to $E$}
    \FOR{each batch $\mathcal{B} \subset \mathcal{D}_{\text{train}}$} 
        \STATE Initialize $\mathcal{L}_{\text{batch}} \leftarrow 0$
        \FOR{each $(x^{(i)}, y^{(i)})$ in $\mathcal{B}$}
            \FOR{each $\theta_n$}
                \STATE Query $\mathcal{T}$ with $(x^{(i)}, \theta_n)$ to obtain $p(y=c \mid \theta_n, x^{(i)})$
            \ENDFOR
            \STATE Pass $(x^{(i)}, Q_{\text{best}})$ into $\mathcal{S}$ to get student logits
            \STATE Compute loss $\mathcal{L}^{(i)} $ (Softmax or Dirichlet)
            \STATE Accumulate: $\mathcal{L}_{\text{batch}} \leftarrow \mathcal{L}_{\text{batch}} + \mathcal{L}^{(i)}$
        \ENDFOR
        \STATE Update student weights $\{A_\ell, B_\ell\}$ in all LoRA layers $\{\ell\}$ using gradient descent
    \ENDFOR
    \STATE Break if $\text{NLL}$ (on training data) has not improved in the recent 5 epochs
\ENDFOR
\STATE Restore the checkpoint for the best $\text{NLL}$
\RETURN Fine-tuned student model $\mathcal{S}^{*}$
\end{algorithmic}
\label{alg:dirichlet-wmle}
\end{algorithm}

\section{Sampling LLMs as Teachers}
\label{sec:sampling_teachers}

We consider two distinct approaches for modeling the predictive uncertainty in LLMs. The first approach is the standard Bayesian learning with uncertainty modeled directly for the model weights. The second approach, termed {Bayesian Prompt Ensembles} (BayesPE), relies on averaging over input prompts, e.g., modeling uncertainty arising from prompt-based variation. 

\subsection{Bayesian Neural Networks}
BNNs model uncertainty by treating the model weights as probability distributions rather than 
deterministic point estimates~\cite{ritter2018scalable,blundell2015weight,neal1996bayesian}. They aim to infer a distribution over model weights, conditioned on the observed data as formalized by the Bayes theorem:
$
q(\theta) := p(\theta \mid \mathcal{D}) = \frac{p(\mathcal{D} \mid \theta)  p(\theta)}{p(\mathcal{D})},
$
where \(\theta\) denotes the model weights and \(\mathcal{D}\) is the training dataset. The resulting posterior distribution encapsulates epistemic uncertainty, which arises from limited knowledge about the model's parameters.
Data uncertainty quantification is achieved by evaluating the predictive distribution, defined formally as:
\begin{equation}
    p(y \mid x, \mathcal{D}) = \int p(y \mid x, \theta)  q(\theta)  d\theta,
    \label{eq:predictive_distribution}
\end{equation}
where \(y\) denotes the prediction for input \(x\), and \(q(\theta)\) is a distribution that encodes uncertainty over some latent parameters \(\theta\).
In practice, the integral in Eq.~\eqref{eq:predictive_distribution} is analytically intractable for complex models such as LLMs. Consequently, it is typically approximated via Monte Carlo (MC) sampling (see Eq.~1).

\subsection{Bayesian Prompt Ensembles} 

BayesPE~\cite{ref1-bayespe} is an alternative black-box method designed to estimate uncertainty in LLMs using ensembles. Ensembles aggregate predictions from several independently trained deterministic models.
The central idea behind BayesPE is to assess the variability of a model's output across multiple semantically equivalent prompts, interpreting this variability as a measure of epistemic uncertainty. Formally, let $\mathcal{A} = \{\theta_1, \theta_2, \dots, \theta_N\}$ denote a set of semantically equivalent prompts. Then, a discrete probability distribution $q(\theta)$ is defined over this prompt set as $q(\theta) = \delta_{\theta \in \mathcal{A}}$, and a weight $w_i$ is assigned  to each prompt $\theta_i$. These weights represent the relative importance or reliability of each prompt concerning the task at hand.

Unlike conventional fine-tuning techniques, BayesPE neither requires a dedicated training dataset nor updates the internal parameters of the language model. Instead, it focuses on \emph{learning prompt weights} using a small labeled validation set. Given such a validation dataset $\mathcal{D}_{\text{val}} = \{(x_j, y_j)\}_{j=1}^M$, 
where $x_j$ represents the $j$-th input example and $y_j$ its corresponding ground-truth label, 
the objective is to infer optimal prompt weights $\{w_i\}_{i=1}^N$ via variational inference.

The BayesPE objective is:
\begin{equation*}    
\label{eq:bayespe_objective}
\mathcal{L}_{\text{BayesPE}} = \sum_{j=1}^M \left( \sum_{i=1}^N w_i \log p(y_j \mid \theta_i, x_j) -\sum_{i=1}^N w_i \log w_i \right),
\end{equation*}
where the first term maximizes the likelihood of correct predictions under each prompt and the second term serves as an entropy regularizer to avoid overconfidence in any single prompt. A higher weight implies that the corresponding prompt consistently yields more accurate or confident predictions on the validation data.

\section{Teachers Comparison: BayesPE vs. Laplace}
\label{sec:teachers}

\begin{figure}[ht]
    \centering
    \includegraphics[width=1.0\linewidth]{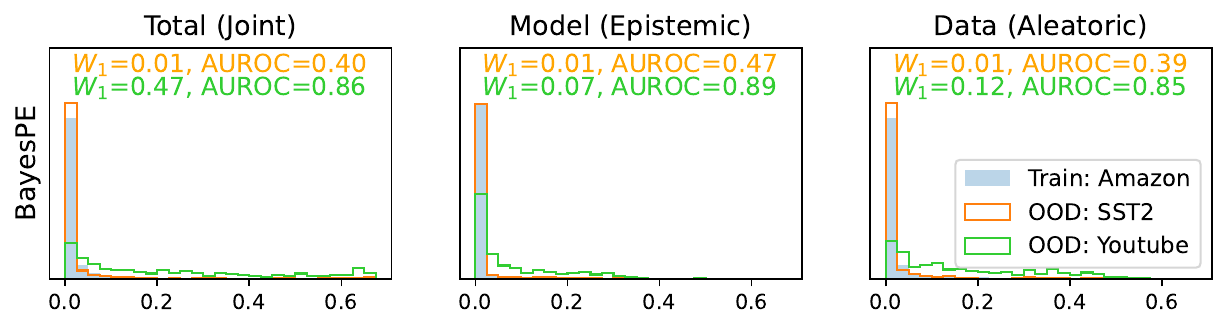}
    \includegraphics[width=1.0\linewidth]{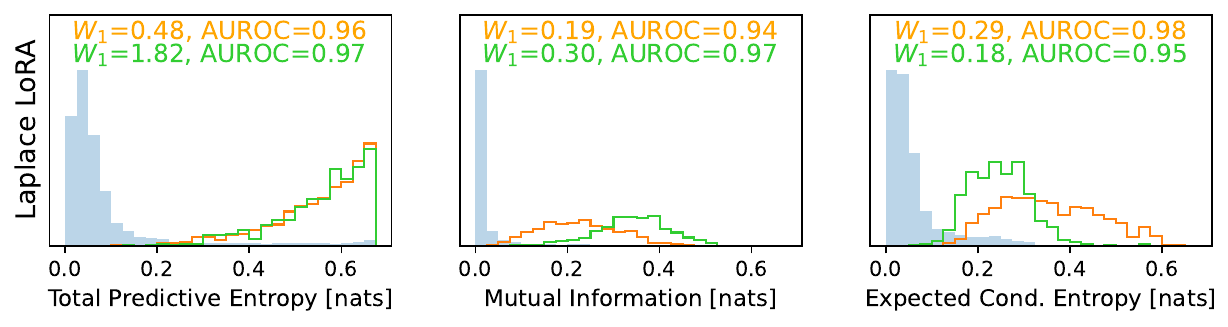}
    \caption{
    Predictive uncertainty distributions for teachers. Each row shows the empirical distributions of \textbf{(left):} total predictive entropy, \textbf{(middle):} mutual information (epistemic uncertainty), and \textbf{(right):} expected conditional entropy (aleatoric uncertainty) on in-domain Amazon reviews (blue) and two OOD datasets: SST2 (orange) and YouTube (green). Wasserstein-1 distance ($W_1$) and AUROC for OOD detection (Amazon vs. YouTube/SST2) are reported.
    }
    \label{fig:ood_teachers}
\end{figure}

We utilize two distinct types of teachers with different characteristics and performance. Table~\ref{tab:teachers}  compares them using Accuracy, ECE, NLL, and Brier score. The Laplace teacher performs noticeably better. In particular, for the Yahoo Answers dataset, BayesPE lags behind. 
Then,
Figure~\ref{fig:ood_teachers} illustrates the predictive distributions and their decompositions for both teachers. Again, the Laplace LoRA teacher performs better in OOD detection, especially for the SST2 dataset. 
Overall, both teachers are comparable on the Amazon and YouTube datasets but differ on the remaining (SST2 and Yahoo) datasets. 
However, it is beyond the scope of this work to further investigate the reasons for this discrepancy 
or try to optimize their training process or hyperparameters. 
In principle, teachers in the presented approach could be any black-box sampling models.

\begin{table}[t!]
    \centering

\caption{
Test set performance of the teacher models. 
}
\label{tab:teachers}

    \begin{tabular}{lllllll}
    \toprule
     Dataset & Model  &  & Accuracy & ECE & NLL & Brier \\
    \midrule
    \multirow[t]{5}{*}{Amazon Reviews} & BayesPE & & {0.959}& 0.021& 0.160& {0.035}\\
    \multirow[t]{3}{*}{} & Laplace & & 0.973 & 0.009 & 0.092 & 0.023 \\
    \cline{1-7}
    \multirow[t]{5}{*}{SST2} & BayesPE & & {0.955}& 0.029& 0.165& {0.037}\\
    \multirow[t]{3}{*}{} & Laplace & & 0.961 & 0.020 & 0.125 & 0.033 \\
    \cline{1-7}
    \multirow[t]{5}{*}{YouTube Comments} & BayesPE & & 0.875& 0.031& 0.295& 0.091\\
    \multirow[t]{3}{*}{} & Laplace & & 0.932 & 0.015 & 0.170 & 0.050 \\
    \cline{1-7}
    \multirow[t]{5}{*}{Yahoo Answers} & BayesPE & & 0.593& 0.194& 2.173& 0.061\\
    \multirow[t]{3}{*}{} & Laplace & & 0.717 & 0.022 & 0.823 & 0.040 \\
    \cline{1-7}
    \bottomrule
    \end{tabular}

\end{table}

\section{Regularization Impact}
\label{sec:regularization_impact}

The concentration parameter $\alpha_0$ of the Dirichlet distribution serves as a critical regularizer that directly influences the model's uncertainty estimation. Higher values of $\alpha_0$ lead to more concentrated probability mass, resulting in peaked distributions that reflect higher confidence. Lower values produce flatter distributions, leading to greater predictive uncertainty.

\begin{figure}[t!]
    \centering

    \includegraphics[width=0.385\linewidth]{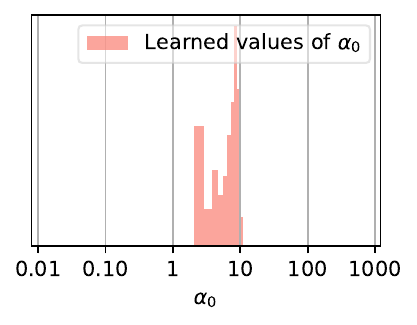}
    \\ (a) Distribution of learned $\alpha_0$-s   
    
    \includegraphics[width=1.0\linewidth]{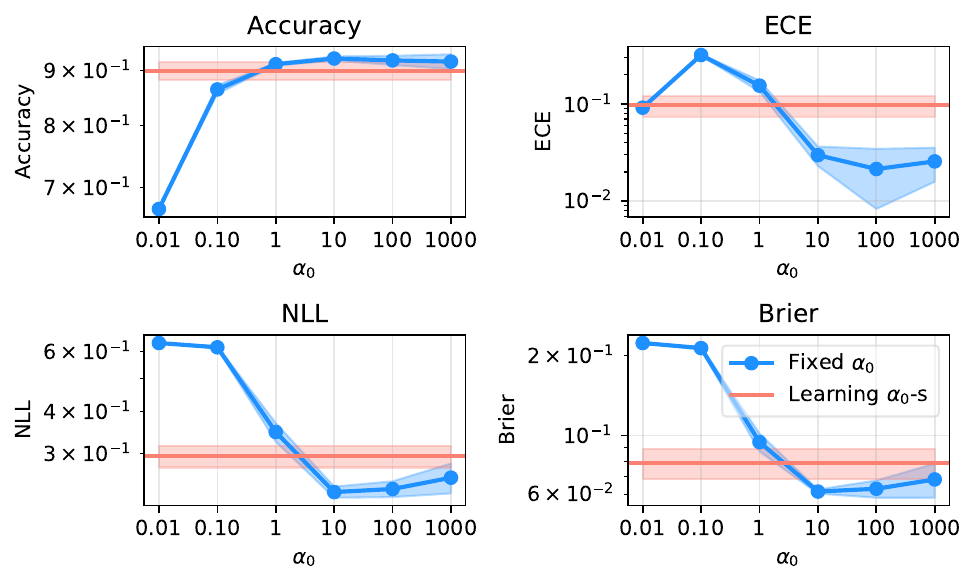}
    \\ (b) Performance comparison (mean $\pm$ std)
  
    \caption{
        (a) Histogram of learned $\alpha_0$ values on the YouTube test set.  
        (b) Comparison between the model with fixed global $\alpha_0$ (red) and the model with a learned, sample-specific $\alpha_0$-s for each test sample (blue), evaluated on the YouTube test set.
    }
    \label{fig:fixed_alphas}
\end{figure}

We evaluate the performance of several regularization strategies for the Dirichlet concentration parameter $\alpha_0$, comparing them against the standard $\text{Dir}(\mathbf{1}_K)$ prior (see Section~3.4) and the non-regularized baseline (the default Dirichlet student). 
Specifically, 
we investigate a 'matching'case where the precision is fixed (e.g., re-normalized at each step to match the desired value) to the number of classes (i.e., $\alpha_0 = K$) and a 'concentrated'case (i.e., $\alpha_0 = 5K$) which provides a larger evidence budget to favor more concentrated distributions.

The results in Table~5 show that the {'no regularization'} approach already demonstrates strong performance and works the best for Amazon and Yahoo datasets.
Across all datasets, the  $\text{Dir}(\mathbf{1}_K)$  prior and the 'matching'precision ($\alpha_0 = K$) strategy consistently leads to the highest ECE and NLL. This suggests that forcing the distribution toward a uniform prior or a low-evidence scale overly penalizes the model's confidence. In contrast, the 'concentrated'fixed precision ($\alpha_0 = 5K$) sometimes yields the best calibration, suggesting that a higher evidence "budget" better matches the teacher's certainty.
We observe such the effect for the SST2 and YouTube datasets, where this regularizer wins in all metrics.
Finally, the proposed $\bar{\alpha}_0$ regularization, which pulls the precision toward a learnable dataset-wide mean, offers a balance between flexibility and stability. While it does not win for any dataset, it  maintains competitive performance overall.

\label{sec:fixed_alpha}

Following the observation 
that the 'concentrated'regularizer often improves performance, we further investigate its impact on the Dirichlet student distilled from the BayesPE teacher. In particular, Figure~\ref{fig:fixed_alphas} illustrates the effect of using a fixed, predefined, global (the same for all inputs) $\alpha_0$ compared to a learned (i.e., unconstrained during optimization) evaluated on the YouTube test set. As shown in Table~2, this is the only dataset for which the Dirichlet student struggles to quantify uncertainty, motivating the exploration of strategies to further enhance performance.

Panel (a) displays the distribution of individually learned $\alpha_0$ values, highlighting their limited variability across samples; for instance, all values fall within the range of 2 to 12. Panel (b) presents a direct performance comparison.

From Panel (b), we observe that varying the global $\alpha_0$ has a substantial impact on all key performance metrics. There exist globally optimal values of $\alpha_0$ for specific metrics (approximately $10$ for Accuracy, NLL, and Brier; around $100$ for ECE). These optimal values are, to a large extent, aligned with the upper end of the range of $\alpha_0$ values learned by the adaptive model.

The results confirm that, while learning $\alpha_0$ per sample achieves consistently strong calibration and likelihood, it is sometimes possible to slightly improve upon these results by selecting an appropriate fixed global $\alpha_0$. 
However, determining this universally optimal global value in a principled and general manner remains an open question. 

\section{Datasets}
For the experiments,
we used the following datasets\footnote{Links to the datasets are provided in our repository.}:
\begin{itemize}
    \item {Amazon Reviews Polarity}~\cite{ref6-He_2016} (Train: 10,000; Test: 5,000): 2 classes
    \item {SST2 (Stanford Sentiment Treebank)}~\cite{ref7-socher-etal-2013-recursive} (Train: 10,000; Test: 872): 2 classes
    \item {Yahoo Answers}~\cite{zhang2015character} (Train: 10,000; Test: 5,000): 10 classes
    \item {YouTube Comments}~\cite{ref9-inproceedings} (Train: 1,100; Test: 711): 2 classes
\end{itemize}
We have selected these datasets for our experiments based on other related works in uncertainty quantification in LLMs such as \cite{ref1-bayespe}. 

\section{Experimental Details}

We performed multiple training runs for the student models using different random seeds (0, 1, 2) to ensure varied LoRA weight initializations. Results are reported as the mean $\pm$ standard deviation across these runs. Training was conducted with a learning rate of $1 \times 10^{-5}$ and a batch size of 16 for the Amazon Reviews, SST2, and YouTube Comments datasets, while a batch size of 1 was used for Yahoo Answers. For test evaluation, a batch size of 16 was applied across all datasets. 

In accordance with prior work on uncertainty modeling~\cite{ref1-bayespe,ICLR2024_07c256a1}, we evaluated performance using Accuracy, Expected Calibration Error (ECE), Negative Log-Likelihood (NLL), Brier Score, and Mean Predictive Entropy. 

Experiments were conducted on two hardware configurations running Ubuntu 22.04 LTS: an NVIDIA A100 (40GB VRAM) for the Amazon, SST2, and YouTube datasets, and an NVIDIA RTX A6000 (49GB VRAM) for the Yahoo Answers dataset.



\section{Limitations and Future Work}

 This study is limited to classification with discrete labels, and does not address open-vocabulary generation or structured prediction. While the Dirichlet output layer allows for explicit separation of epistemic and aleatoric uncertainty, it can still underestimate epistemic uncertainty on some datasets, indicating the potential benefit of hierarchical evidential priors or hybrid Bayesian-evidential models. The experiments are conducted with a 7-billion parameter backbone, so outcomes may differ when scaling to larger or sparse architectures.

Possible future directions include generalizing the approach to regression and sequence-to-sequence tasks, integrating retrieval-augmented or instruction-tuned teachers, exploring task-adaptive prompt selection for improved uncertainty estimation, and evaluating practical gains in decision-making applications such as clinical triage or financial risk assessment.

\end{document}